\documentclass[runningheads]{llncs}

\usepackage{eccv}

\usepackage{eccvabbrv}

\usepackage{graphicx}
\usepackage{booktabs}

\usepackage{multirow}      
\usepackage{makecell}      
\usepackage{siunitx}       
\usepackage[table]{xcolor} 
\usepackage{wrapfig}
\usepackage[a4paper,width=122mm,left=12mm,paperwidth=146mm,height=190mm,top=12mm,paperheight=217mm]{geometry}

\usepackage[accsupp]{axessibility}  

\usepackage{hyperref}

\usepackage{orcidlink}

\definecolor{catgray}{gray}{0.9}  %

\begin{document}

\title{Anchor Forcing: Anchor Memory and Tri-Region RoPE for Interactive Streaming Video Diffusion} 

\titlerunning{Anchor Forcing}

\author{Yang Yang$^{1,2}$ ~~
Tianyi Zhang$^{2}$ ~~
Wei Huang$^{3}$ ~~
Jinwei Chen$^{2}$ ~~
Boxi Wu$^{{1}}$\thanks{Corresponding authors.} \\
Xiaofei He$^1$ ~~
Deng Cai$^1$ ~~
Bo Li$^2$ ~~
Peng-Tao Jiang$^{2\star}$ \vspace{1pt}\\
}

\authorrunning{Y.~Yang et al.}

\institute{$^1$Zhejiang University ~~ $^2$vivo BlueImage Lab ~~ $^3$University of Hong Kong
\url{https://github.com/vivoCameraResearch/Anchor-Forcing}
}

\maketitle

\begin{figure}[th]
    \centering
    \includegraphics[width=1.0\linewidth]{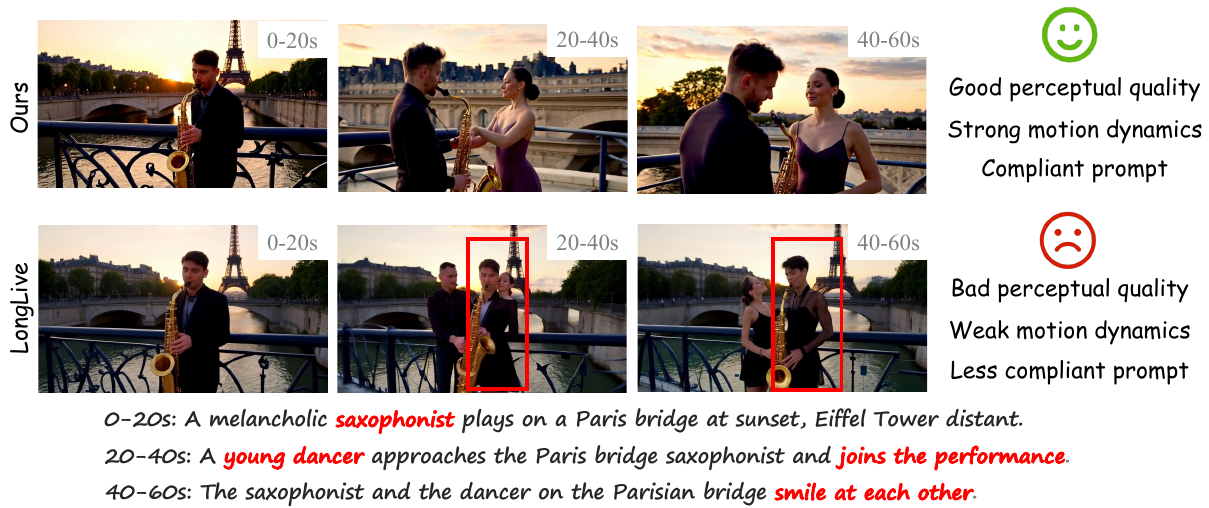}
    \caption{We propose \textbf{Anchor Forcing}, which supports prompt switches that introduce new subjects and actions while preserving context, motion quality, and temporal coherence across clips. In contrast, prior methods degrade over time and often fail to realize newly introduced interactions, as highlighted by the red boxes. Red text denotes the interaction newly specified in each segment.}
    \label{fig:motivation}
\end{figure}

\begin{abstract}
Interactive long video generation requires prompt switching to introduce new subjects or events, while maintaining perceptual fidelity and coherent motion over extended horizons.
Recent distilled streaming video diffusion models reuse a rolling KV cache for long-range generation, enabling prompt-switch interaction through re-cache at each switch.
However, existing streaming methods still exhibit progressive quality degradation and weakened motion dynamics.
We identify two failure modes specific to interactive streaming generation:
(i) at each prompt switch, current cache maintenance cannot simultaneously retain KV-based semantic context and recent latent cues, resulting in weak boundary conditioning and reduced perceptual quality; and
(ii) during distillation, unbounded time indexing induces a positional distribution shift from the pretrained backbone's bounded RoPE regime, weakening pretrained motion priors and long-horizon motion retention.
To address these issues, we propose \textbf{Anchor Forcing}, a cache-centric framework with two designs.
First, an anchor-guided re-cache mechanism stores KV states in anchor caches and warm-starts re-cache from these anchors at each prompt switch, reducing post-switch evidence loss and stabilizing perceptual quality.
Second, a tri-region RoPE with region-specific reference origins, together with RoPE re-alignment distillation, reconciles unbounded streaming indices with the pretrained RoPE regime to better retain motion priors.
Experiments on long videos show that our method improves perceptual quality and motion metrics over prior streaming baselines in interactive settings.

\keywords{Interactive Video Generation \and Quality Preservation \and Motion Enhancement}
\end{abstract}

\section{Introduction}
\label{sec:intro}

Interactive long video generation is becoming increasingly important for creative content creation~\cite{po2025long,yu2025context} and film-level scene development~\cite{cai2025mixture,li2025svi}. Compared with short clips, long-horizon generation must sustain narrative coherence while continuously producing visually faithful frames and plausible temporal dynamics. A straightforward approach~\cite{po2025long,yu2025context,cai2025mixture,li2025svi,li2025magicworld} that applies dense spatiotemporal attention over all frames is computationally prohibitive at long horizons, which makes high-quality generation difficult to deploy in interactive scenarios where users expect low latency and frequent edits.

To improve practicality, recent distilled streaming video diffusion models~\cite{yang2025longlive,yesiltepe2025infinity,ji2025memflow,yi2025deep} adopt causal temporal attention and reuse a rolling transformer KV cache. With causal attention~\cite{yin2025causvid}, each denoising step conditions only on previously generated frames, while cached keys and values enable efficient reuse of long-range context across steps without recomputing attention over the entire history. In practice, these systems typically combine short-window attention with a small set of sink tokens~\cite{xiao2023efficient} to stabilize long-horizon generation and support prompt-switch interaction by updating the cache state at switch boundaries. 

Despite this progress, existing approaches largely emphasize extending temporal length, while interactive long video generation still exhibits two recurring failure modes. 
As shown in \cref{fig:motivation}, cache maintenance at prompt switches often fails to preserve relevant semantic evidence, weakening cross-segment continuity in interactive videos and hurting perceptual quality. In particular, re-cache-based methods~\cite{yang2025longlive,ji2025memflow} rebuild the cache by discarding prior semantic KV states from earlier stages, which reduces the available semantic context for post-switch conditioning and degrades perceptual quality in interactive long videos.
Second, during distillation, streaming generation~\cite{yang2025longlive,ji2025memflow} uses RoPE indices that grow unbounded over time, while the pretrained backbone is optimized under a finite and bounded positional range. This gap induces a positional distribution shift, weakening the transfer and retention of pretrained motion priors.
Together, these effects make interactive long videos prone to degraded visual quality and weakened motion dynamics.

To bridge these gaps, we propose Anchor Forcing, a cache-centric framework for interactive long video generation under streaming constraints, designed to preserve visual quality and retain motion dynamics.
Our framework has two core designs.
First, we introduce an anchor-guided re-cache mechanism. At each prompt switch, we generate a short initial segment conditioned on the new prompt and store its key–value states as a junction cache. We then build an anchor memory by integrating the junction cache with the sink cache collected at the early stage of generation and the local cache that preserves the most recent context. This anchor memory is then used to warm-start re-caching for subsequent generations under the new prompt, reducing evidence loss and stabilizing post-switch quality.
Second, we propose a tri-region RoPE with region-specific reference origins and indexing. Instead of using an unbounded positional index, we adopt a relative, range-bounded indexing scheme that keeps effective RoPE positions within the bounded regime observed during pretraining. Moreover, we treat the sink, junction, and local caches separately and assign each region its own RoPE encoding. During distillation, the generator adaptively attends across regions and leverages complementary cues, improving motion-prior retention and long-horizon dynamics.

Empirically, Anchor Forcing improves perceptual quality and motion dynamics on VBench benchmarks. In summary, our contributions are:
\begin{itemize}
    \item We introduce an anchor cache that stores anchor memory after each prompt switch, enabling anchor-guided re-cache to reduce post-switch evidence loss and stabilize generation quality.
    \item We propose a tri-region RoPE that partitions the anchor memory into three regions, keeps effective RoPE positions within the bounded range seen in pretraining, and encourages adaptive cross-region feature retrieval during distillation, thereby improving long-horizon motion-prior retention.
    \item We achieve state-of-the-art results on VBench in interactive settings on perceptual quality and motion-related metrics, while consistently outperforming prior streaming baselines overall.
\end{itemize}

\section{Related Work}


\subsection{Diffusion-Based Long Video Generation}

A straightforward route to longer videos is to scale the context length of video diffusion Transformers and train models to utilize large spatiotemporal windows~\cite{li2025svi, he2022latent, cai2025mixture,villegas2022phenaki, guo2025long, dalal2025one, chen2023seine, po2025long, qiu2023freenoise, lu2025freelong++}.
For example, SVI~\cite{li2025svi} enables infinite-length video generation by error-recycling fine-tuning that trains the model to detect and correct its own accumulated errors.
FreeNoise~\cite{qiu2023freenoise} adopts noise rescheduling with window-based temporal attention.
LVDM~\cite{he2022latent} performs hierarchical video synthesis in a compact 3D latent space to improve efficiency.
Mixture of Contexts~\cite{cai2025mixture} learns sparse routing to a few informative segments plus mandatory anchors, improving efficiency compared to uniform long-context attention.
FreeLong~\cite{lu2024freelong} improves long-range consistency by blending temporal frequency components at inference, and FreeLong++~\cite{lu2024freelong} further extends this idea with multi-band spectral fusion to integrate multi-frequency temporal signals.
Seine~\cite{chen2023seine} uses random-mask conditioned video diffusion to extend shot-level clips into story-level videos with text-guided transitions.
SSM~\cite{po2025long} investigates state-space based architectures to extend temporal memory without quadratic attention growth.
Dalal et al.~\cite{dalal2025one} augment a pretrained diffusion transformer with test-time training (TTT) layers, maintaining long-horizon consistency while keeping self-attention computationally tractable.
LaVie~\cite{wang2025lavie} adopts a cascaded generation pipeline to enhance long-range coherence.
LCT~\cite{guo2025long} expands pretrained single-shot video diffusion models to multi-shot scenes by enlarging the context window and extending full attention across shots.
Despite these advances, bidirectional processing and repeated denoising steps still pose practical bottlenecks for streaming and prompt-switchable generation.

\subsection{Autoregressive-Based Long Video Generation}
To enable real-time or long-horizon streaming, a growing line of work~\cite{yin2025causvid, huang2025self, yesiltepe2025infinity, yang2025longlive, yi2025deep, ji2025memflow, cui2026lol, henschel2025streamingt2v, teng2025magi,jin2024pyramidal, chen2024diffusion, song2025history, feng2024matrix, gu2025long,villegas2022phenaki,yin2023nuwa} adopts causal temporal modeling and reuses temporal KV caches.
CausVid~\cite{yin2025causvid} reframes video diffusion as a causal generation process and reduces the number of inference steps via distribution-matching distillation.~\cite{yin2024dmd}.
Self-Forcing~\cite{huang2025self} reduces the train–test mismatch in autoregressive video diffusion by simulating KV caching during training, thereby mitigating exposure bias.
Infinity-RoPE~\cite{yesiltepe2025infinity} studies position-encoding and cache behaviors under long rollouts and proposes KV flush to improve horizon and responsiveness during interaction.
LongLive~\cite{yang2025longlive} supports prompt switching in long rollouts via frame-level autoregressive generation with short-window causal attention, frame-sink tokens, and a KV-recache update scheme.
MemFlow~\cite{ji2025memflow} enables narrative prompting by using prompt-conditioned memory retrieval with sparse activation over a memory bank.
Nevertheless, streaming methods still exhibit long-horizon failure modes, including drift and regressions toward early content. LoL~\cite{cui2026lol}analyzes sink-collapse behavior and proposes a training-free mitigation to reduce such degeneracy at extreme horizons.
LongVie~\cite{gao2025longvie} introduces multimodal guidance, unified noise initialization, and degradation-aware training, and StreamingT2V~\cite{henschel2025streamingt2v} extends these ideas with short- and long-term memory modules for coherent text-to-video generation.
SkyReels-V2~\cite{chen2025skyreels} combines diffusion forcing with a film-structure planner and multimodal controls. FramePack~\cite{zhang2025frame} compresses input frames into a fixed-size context to address memory and efficiency constraints, and FAR~\cite{gu2025long} further improves autoregressive generation by coupling a high-resolution short-term context with a compressed long-term context via flexible positional encoding.
History-guided video diffusion~\cite{song2025history} incorporates flexible-length historical context to strengthen temporal consistency over extended rollouts, while Pyramidal-flow~\cite{jin2024pyramidal} proposes a multi-scale flow-matching design to reduce computation.
%

Overall, existing autoregressive streaming approaches substantially improve efficiency and temporal extent. Yet, two issues remain: cache maintenance at prompt switches fails to jointly retain KV-based semantic context and recent latent cues, weakening boundary conditioning and perceptual quality, and unbounded time indexing during distillation deviates from the pretrained backbone’s bounded RoPE regime, undermining motion-prior transfer and long-horizon motion retention.

\section{Method}
\subsection{Preliminaries}
\textbf{Streaming autoregressive video diffusion.}
We consider streaming autoregressive (AR) video diffusion that generates long videos in small temporal chunks while reusing a temporal Key--Value (KV) cache to reduce per-step computation~\cite{yin2025causvid}.
At each step, the model conditions on (i) a local cache that stores recent latent tokens to capture short-term dynamics, (ii) a small global sink cache that provides a persistent surrogate of long-range context, and (iii) the current text prompt embedding.
Practical systems~\cite{yang2025longlive,ji2025memflow} typically combine short-window causal attention with a bounded KV cache composed of local and sink caches, enabling long-horizon generation with bounded attention context. Moreover, they often employ distribution matching distillation~\cite{yin2024dmd} to distill multi-step diffusion models into a few-step generator.
Given a noised latent $x_t$ at diffusion timestep $t$ and prompt embedding $c$, the model predicts a denoising direction conditioned on the assembled memory:
\begin{equation}
    \hat{v}_{\theta} = f_{\theta}(x_t,\, t,\, c;\, \mathcal{M}),
\end{equation}
where $\mathcal{M}$ denotes the memory constructed from the sink cache and the rolling local cache, and is updated in a rolling manner after each generated chunk.
Interaction is realized by updating the conditioning prompt at interaction boundaries during inference (including prompt switches and edits), requiring the model to incorporate updated instructions while maintaining temporal coherence.

\noindent \textbf{Distribution Matching Distillation.}
Let $G_\theta$ denote the few-step generator parameterized by $\theta$, which maps Gaussian noise $\epsilon \sim \mathcal{N}(0,I)$ to a generated sample $x_0=G_\theta(\epsilon)$. To distill a multi-step diffusion model into $G_\theta$, the distribution matching distillation~\cite{yin2024dmd,yin2024onestep} (DMD) minimizes the reverse KL divergence between the real data distribution $p_{\text{data}}$ and the output distribution of the generator $p_{\text{gen}}$. The gradient of the reverse KL divergence can be approximated as the difference between two score functions:
\begin{equation}
\begin{aligned}
\nabla_\theta \mathcal{L}_{\text{DMD}} &= \mathbb{E}_{t}\left[ \nabla_\theta \text{KL}(p_{\text{gen}}(x_t)||p_{\text{data}}(x_t)) \right] \approx \mathbb{E}_{x_t,t}\left[ \left( s_{\text{gen}}(x_t) - s_{\text{data}}(x_t) \right) \frac{\partial x_t}{\partial \theta} \right],
\end{aligned}
\label{eq:eq2}
\end{equation}
where $x_t=\alpha_tx_0+\beta_t\epsilon'$ denotes the re-noised sample of $x_0$ using the noise schedule coefficients $\alpha_t$ and $\beta_t$ with $\epsilon'\sim \mathcal{N}(0,I)$, $s_{\text{data}}$ and $s_{\text{gen}}$ represent the score functions of the real data and generator’s output distribution, respectively. 
During training, DMD uses a fixed pre-trained teacher diffusion model to approximate the score function of the real data distribution, and simultaneously trains a neural network to estimate the score function of the generator’s output distribution. Finally, the gradient derived in Eq.~\ref{eq:eq2} is used to guide the generator to align its output distribution with the data distribution.

\begin{figure}[t]
    \centering
    \includegraphics[width=1.0\linewidth]{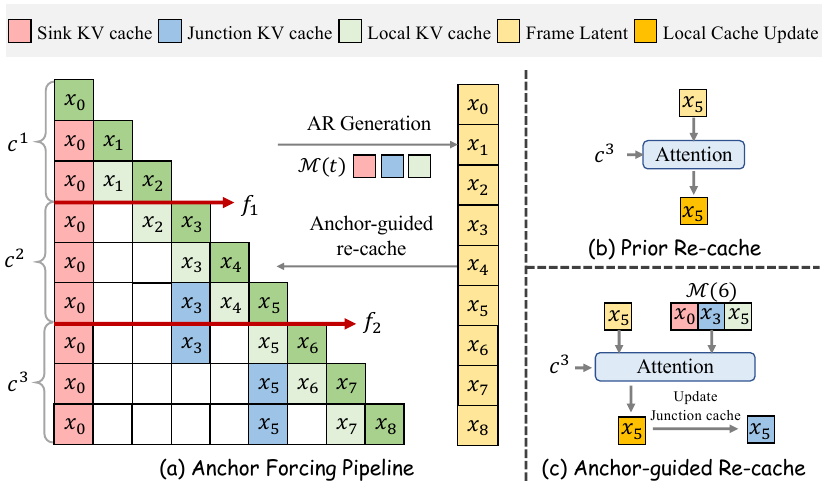}
    \caption{\textbf{The Anchor Forcing pipeline.}
    (a) \textbf{Overview of Anchor Forcing} in an interactive setting with two prompt switches. We denote the anchor memory for generating frame $t$ as $\mathcal{M}(t)$, and apply anchor-guided re-cache at $f_1$ and $f_2$ to update the local KV cache under the new prompt condition.
    (b) \textbf{Prior re-cache~\cite{yang2025longlive}}. It rebuilds the local cache solely from historical frame latents, which fails to retain prior KV evidence across prompt switches.
    (c) \textbf{Anchor-guided re-cache} at $f_2$. It augments re-cache with the anchor memory $\mathcal{M}(6)$ and refreshes the junction caches $x_5$.}
    \label{fig:method}
\end{figure}

\subsection{Anchor-guided Re-cache Mechanism}


\textbf{Problem formulation}. 
In interactive text-to-video generation, the conditioning prompt is updated at interaction boundaries, and a cache maintenance operation is typically invoked to rebuild the KV cache for the new prompt condition.
As illustrated in \cref{fig:method}(b). 
Specifically, re-cache-based methods~\cite{yang2025longlive,ji2025memflow} rebuild the cache by evicting prior KV states, which removes informative context from the preceding stage and weakens cross-switch semantic continuity.
Overall, these deficiencies reduce the availability of effective conditioning cues around interaction boundaries, leading to degraded video quality in interactive long videos.

\noindent\textbf{Anchor Memory}. 
We maintain an anchor memory $\mathcal{M}(t)$ composed of three cache parts. The sink cache $\mathcal{S}$ stores KV states from the earliest $N_S$ generated frames and remains fixed throughout generation, providing global semantic context. The junction cache stores boundary-adjacent evidence captured immediately after each prompt switch, preserving newly introduced interaction cues. The local cache retains the most recent short-range context to support fine-grained continuity.
We consider a prompt stream $\mathcal{C}=\{c^{1},c^{2},\dots,c^{m}\}$ with interaction boundaries $\mathcal{F}=\{f_{1},f_{2},\dots,f_{m-1}\}$, where $f_i$ denotes the frame index at which the prompt updates to $c^{i}$. 
For any frame $t$, we define:
\begin{equation}
\pi(t)=\max\{\, i \mid f_i \le t,\ i\in[1,m-1] \,\},
\qquad
s(t)=f_{\pi(t)}.
\end{equation}
The active prompt at frame $t$ is $c^{\pi(t)}$, and $s(t)$ is the most recent frame index of the interaction boundary.
We denote by $\mathbf{KV}[a:b]$ the per-layer KV state of frames $a$ to $b$ for causal Transformer layers~\cite{yin2025causvid}.
After each boundary at $f_i$, we take the first $N_J$ frames under the updated prompt and store their per-layer KV states in the  caches:
\begin{equation}
\mathcal{J}_{i}=\mathbf{KV}\!\left[f_i : f_i+N_J-1\right].
\end{equation}
The junction caches are refreshed at each interaction boundary and kept fixed until the next boundary. 
In contrast, the rolling local KV cache maintains only a window of length $W$:
\begin{equation}
\mathcal{M}_{\ell}(t)=\mathbf{KV}\!\left[t-W+1 : t\right].
\end{equation}
For simplicity, as shown in \cref{fig:method}(a), we illustrate the rolling update of the anchor memory at the two prompt switches, $f_1$ and $f_2$, under the setting $N_J = W = 1$.
To avoid redundancy with local KV cache, junction cache are activated only after the junction frames have left the local window. Accordingly, the anchor memory $\mathcal{M}(t)$ used to generate frame $t$ is
\begin{equation}
\mathcal{M}(t)=\big[\mathcal{S}\ ;\ \mathcal{M}_{\ell}(t)\ ;\ \delta_t\,\mathcal{J}_{\pi(t)}\big],\;
\text{where}\;
\delta_t=\mathbb{I}\!\left[s(t)+N_J \le t-W\right].
\end{equation}
When $\delta_t=0$, the junction frames are still contained in $\mathcal{M}_{\ell}(t)$ and their KV states are directly accessible via the local context. 
When $\delta_t=1$, they have been evicted by rolling updates, and $\mathcal{J}_{\pi(t)}$ provides a persistent substitute that keeps junction evidence reachable and repeatedly attendable in later segments.

\noindent\textbf{Anchor-guided Re-cache}.
In interactive text-to-video generation, the prompt is updated at interaction boundaries, and a re-cache operation is typically invoked to rebuild the cache state under the new prompt condition.
However, as shown in \cref{fig:method}(b), re-cache~\cite{yang2025longlive} discards prior KV states, which reduces the available semantic context for post-switch conditioning and removes boundary-relevant evidence that is crucial for continuity and high-fidelity generation after the switch.
%
%
To mitigate this issue, we propose an anchor-guided re-cache mechanism, as illustrated in \cref{fig:method}(c). Unlike prior re-cache, we recompute the local cache under the updated prompt by additionally leveraging the anchor memory from the previous segment.
Subsequent streaming generation is conditioned on the refreshed local cache together with the sink cache, with the junction cache providing persistent boundary-adjacent evidence to support smooth motion evolution while promptly adapting to the updated prompt.

\subsection{Tri-region RoPE}
\label{sec:rope}

RoPE~\cite{su2024roformer} is widely used to encode temporal positions in Transformer attention.
In streaming generation~\cite{yang2025longlive,ji2025memflow} with rolling KV caches, a straightforward implementation assigns RoPE indices using a global latent frame index that grows monotonically over time.
As the rollout length increases, these indices may exceed the positional range observed during pretraining, inducing a positional distribution shift and weakening the transfer of pretrained motion priors and long-horizon motion retention.

In our setting, the generator is initialized from CausVid~\cite{yin2025causvid}, which is trained with ODE trajectories capped at 21 latent frames (distilled from Wan~\cite{wan2025}); thus, the model is primarily exposed to RoPE indices within a bounded range.
Accordingly, during streaming inference we replace global absolute indexing with relative, range-bounded indexing when attending to cached tokens, and cap the effective RoPE index to $P_{\max}=21$ to stay within the pretrained positional range.
We adopt a relative, range-bounded indexing scheme, termed tri-region RoPE. Specifically, when generating frame $t$, the anchor memory is $\mathcal{M}(t)=[\mathcal{S}\ ;\ \mathcal{M}_{\ell}(t)\ ;\ \delta_t \mathcal{J}_{\pi(t)}]$.
We then assign region-specific, bounded RoPE indices to cached keys from the sink, local, and clip regions as follows.
Sink caches form a small persistent memory with a fixed order.
We use 0 to $N_S-1$ as its index.
Let the local cache cover a rolling window ending at $t$.
When $t < P_{\max}$, we use the latent index as the RoPE index.
When $t \ge P_{\max}$, we re-index keys relative to the $P_{\max}$ and window of length $W$:
\begin{equation}
p^{\mathrm{L}}(k,t)=
\begin{cases}
t, & t < P_{\max},\\
P_{\max}- W + k, & t \ge P_{\max},
\end{cases}
\quad where \quad k \in [0, W-1]
\end{equation}
where $k$ denotes the index of the local caches. At the same time, the query of the current frame $t$ also uses the same encoding form.
When the junction cache $\mathcal{J}_{\pi(t)}$ is activated, we place its indices immediately in front of the local cache:
\begin{equation}
p^{\mathrm{J}}(k,t)=p^{\mathrm{L}}(k,t) - N_J.
\end{equation}

These definitions keep the RoPE indices used in attention bounded by $P_{\max}$, rather than growing with the rollout length. 
Coupled with distillation training, the generator can better exploit the induced relative geometry across cache regions, learning to retrieve complementary semantic cues from different regions, which leads to stronger motion dynamics in long-horizon interactive generation.

\section{Experiments}

\subsection{Implementation Details}
Anchor Forcing is built upon Wan2.1-T2V-1.3B~\cite{wan2025} and generates 5-second videos at $832\times480$ resolution.
We first train the model using 16k ODE sampled from the base model, and initialize it with causal attention masking following CausVid~\cite{yin2025causvid}.
Text prompts are drawn from the filtered and LLM-augmented VidProM~\cite{wang2024vidprom} dataset.
We then follow LongLive~\cite{yang2025longlive} to train with DMD, switch to short-window attention with sink tokens, and perform streaming long-tuning. During streaming long-tuning, each iteration continues the model’s own rollout by generating consecutive 5-second clips up to a maximum length of 60 seconds. In our implementation, the sink cache and anchor cache store KV states for $N_S=3$ and $N_J=3$ latent frames, respectively, and the rolling local cache uses a window size of $W=9$. Optimization uses AdamW for both actor and critic with learning rates $lr = 1.0 \times 10^{-5}$ and $lr_{critic} = 2.0 \times 10^{-6}$. Training takes approximately 24 hours on 56 A800 GPUs.

\subsection{Metrics}
We evaluate our method using VBench~\cite{huang2023vbench} and VBenchLong~\cite{huang2025vbench++}.
For the 60-second interactive setting and all of the ablation study, we evaluate VBenchLong~\cite{huang2025vbench++} dimensions including subject consistency, background consistency, motion smoothness, dynamic degree, aesthetic quality, and imaging quality. 
Following the standard VBench protocol~\cite{huang2023vbench}, these dimensions are normalized and aggregated using the official coefficients to obtain the overall score.
For semantic adherence under prompt interaction, we segment each video at prompt boundaries and compute a clip-wise semantic score using ViCLIP~\cite{wang2022internvideo}.
For the 30-second and 5-second single-prompt settings, we follow Self-Forcing~\cite{huang2025self} by using their rewritten prompts, and evaluate all VBench dimensions.

\begin{table}[t]
\centering
\scriptsize
\caption{Interactive long video evaluation results. The dynamic degree and quality scores are reported on the whole 60s sequence. CLIP scores are reported on 10s video segments with the same semantics ($\uparrow$ higher is better).}
\sisetup{
  table-number-alignment = center,
  table-format = 2.2,
  round-mode = places,
  round-precision = 2,
  detect-weight = true
}
\resizebox{\linewidth}{!}{
\begin{tabular}{l S S  *{6}{S}}
\toprule
\multirow{2}{*}{\textbf{Method}} &
\multicolumn{1}{c}{\multirow{2}{*}{\makecell{Dynamic\\Degree $\uparrow$}}} &
\multicolumn{1}{c}{\multirow{2}{*}{\makecell{Quality\\Score $\uparrow$}}} &
\multicolumn{6}{c}{CLIP Score $\uparrow$} \\
\cmidrule(lr){4-9}
& & & \multicolumn{1}{c}{0--10s} & \multicolumn{1}{c}{10--20s} &
\multicolumn{1}{c}{20--30s} & \multicolumn{1}{c}{30--40s} &
\multicolumn{1}{c}{40--50s} & \multicolumn{1}{c}{50--60s} \\
\midrule
Infinity-RoPE~\cite{yesiltepe2025infinity}  & 35.17 & 79.98 & 23.87 & 22.62 & 22.16 & 21.82 & 22.20 & 22.20 \\
MemFlow~\cite{ji2025memflow}  & 45.47 & 79.35 & \textbf{24.87} & 24.08 & 23.02 & 22.87 & 22.35 & 21.59 \\ 
LongLive~\cite{yang2025longlive} & 26.37 & 78.92 & 24.82 & 24.18 & 23.30 & 22.98 & 23.40 & 22.82 \\
\textbf{Anchor Forcing}   & \textbf{73.00}  & \textbf{82.25} & 24.48 & \textbf{24.35} & \textbf{23.50} & \textbf{23.53} & \textbf{23.61} & \textbf{23.63} \\
\bottomrule
\end{tabular}}
\label{tab:interactive}
\end{table}

\begin{table}[t]
  \setlength{\tabcolsep}{3.5pt} 
  \caption{
    \textbf{Comparison with relevant baselines under 5s setting.} We compare Anchor Forcing with representative open-source models of similar parameter sizes and resolutions. Evaluation scores are calculated on the standard prompt of VBench.
  }
  \label{tab:short}
  \centering
\resizebox{\linewidth}{!}{
\begin{tabular}{lccccccc}
  \toprule
  \multirow{2}{*}{Model} & \multirow{2}{*}{\#Params} & \multirow{2}{*}{Resolution} & \multirow{2}{*}{Dynamic $\uparrow$} & \multicolumn{3}{c}{Evaluation scores $\uparrow$}\\
  \cmidrule(lr){5-7}
   &  &  &  & Total & Quality & Semantic \\
  \midrule
  \rowcolor{catgray}
  \multicolumn{7}{l}{\textit{Real-data-trained models}}\\
  Pyramid Flow~\cite{jin2024pyramidal} & 2B   & $640{\times}384$ & 68.06 & 81.56 & 83.89 & 72.24 \\
  LTX-Video~\cite{HaCohen2024LTXVideo} & 1.9B & $768{\times}512$ & 75.00 & 82.33 & 85.20 & 70.83 \\
  SkyReels-V2~\cite{chen2025skyreels} & 1.3B & $960{\times}540$ & 48.61 & 81.72 & 82.93 & 76.90 \\
  Wan2.1~\cite{wan2025}   & 1.3B & $832{\times}480$ & 73.61  & 84.27 & 85.24 & 80.43 \\
  \midrule
  \rowcolor{catgray}
  \multicolumn{7}{l}{\textit{Streaming AR models}}\\
  CausVid~\cite{yin2025causvid} & 1.3B & $832{\times}480$ & 63.89 & 82.96 & 83.94 & 79.02 \\
  Self-Forcing~\cite{huang2025self} & 1.3B & $832{\times}480$ & 63.89 & 83.83 & 84.60 & 80.79 \\
  Infinity-RoPE~\cite{yesiltepe2025infinity} & 1.3B & $832{\times}480$ & 58.33 & 83.55 & 84.34 & 80.43 \\
  DeepForcing~\cite{yi2025deep} & 1.3B & $832{\times}480$ & 63.89 & 83.85 & 84.61 & 80.84 \\
  MemFlow~\cite{ji2025memflow} & 1.3B & $832{\times}480$ & 50.00  & 81.76 & 83.74 & 73.86 \\
  LongLive~\cite{yang2025longlive} & 1.3B & $832{\times}480$ & 37.50  & 82.81 & 83.26 & \textbf{81.01} \\
  \textbf{Anchor Forcing} & 1.3B & $832{\times}480$ & \textbf{73.61} & \textbf{83.99} & \textbf{84.84} & 80.57 \\
  \bottomrule
\end{tabular}
}
\end{table}

\begin{figure}[t]
    \centering
    \includegraphics[width=1.0\linewidth]{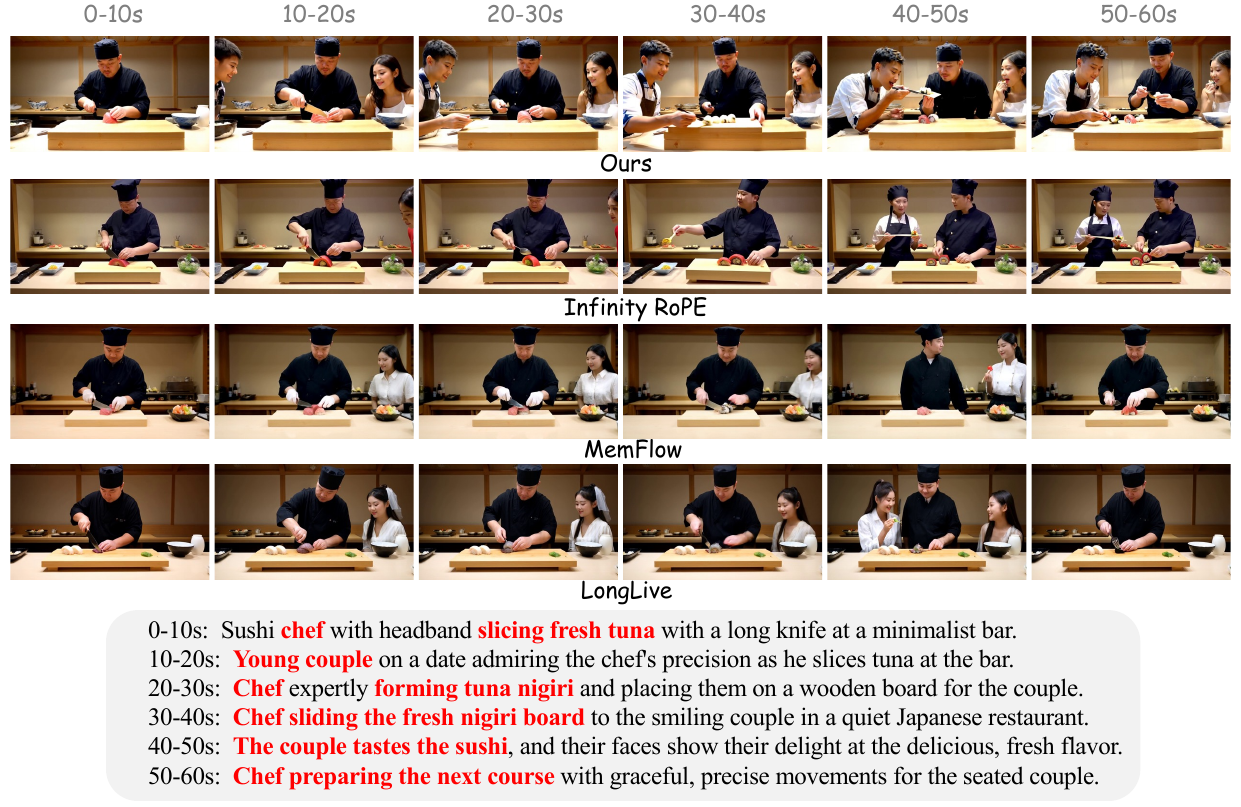}
    \caption{Qualitative comparison on interactive long-video generation. Compared with baselines, Anchor Forcing achieves stronger prompt compliance, more coherent dynamic motion, and higher long-range visual quality. Red text marks the interactive content newly introduced in each segment.}
    \label{fig:interactive compare}
\end{figure}

\subsection{Comparisons for Interactive Generation}
For interactive long-form videos with multiple prompt switches, few prior methods support true streaming generation. We implement this setting for three representative baselines, namely Infinity-RoPE~\cite{yesiltepe2025infinity}, MemFlow~\cite{ji2025memflow}, and LongLive~\cite{yang2025longlive}, and compare our approach against them. For a fair comparison, we follow the interactive benchmark introduced in MemFlow~\cite{ji2025memflow}, which contains 100 narrative scripts. Each script comprises six consecutive 10-second prompts, producing 100 videos with a total duration of 60 seconds.
We evaluate dynamic degree and overall quality on the full 60-second sequence using VBenchLong~\cite{huang2025vbench++}, and report segment-wise semantic adherence by computing ViCLIP~\cite{wang2022internvideo} scores on each 10-second segment with its corresponding prompt.

As shown in~\cref{tab:interactive}, Anchor Forcing achieves the best dynamic degree and quality score among all methods, demonstrating strong motion dynamics and perceptual quality in interactive long videos.
Moreover, while early-segment semantic scores are comparable across methods, Anchor Forcing maintains higher CLIP scores in later segments with substantially less temporal degradation, indicating more stable semantic adherence over long horizons under prompt interaction. 
Additionally, \cref{fig:interactive compare} presents a challenging multi-stage interactive scenario and a qualitative comparison with representative streaming baselines.
Our method follows prompt-induced interactions more faithfully: it introduces the new subjects in the 10--20s segment as specified and responds to subsequent action prompts in the 30--60s segment with more accurate and temporally coherent motion.
Compared with prior methods, our results exhibit stronger motion dynamics, with smoother transitions and more consistent content updates across prompt switches, while maintaining higher visual fidelity throughout the sequence.
In contrast, representative baselines often struggle to follow interactive prompts and sustain strong dynamics: they fail to reliably introduce the ``young couple'' in 10--20s and produce weak or ambiguous actions (e.g., ``slides a board of fresh nigiri'' and ``tastes the sushi'') in 30--60s, whereas Anchor Forcing generates clearer, more plausible prompt-aligned motion.

\begin{table}[t]
\centering
\caption{\textbf{Comparison with relevant baselines under 30s setting.} We compare Anchor Forcing with representative open-source models. Evaluation scores are calculated on the standard prompt of VBench-Long.}
\label{tab:long}
\setlength{\tabcolsep}{6pt}
\renewcommand{\arraystretch}{1.15}
\resizebox{1.0\linewidth}{!}{%
\begin{tabular}{lcccccc}
\toprule
\textbf{Model} & \makecell{Total\\Score $\uparrow$} & \makecell{Quality\\Score $\uparrow$} & \makecell{Semantic\\Score $\uparrow$} & \makecell{Dynamic\\Score $\uparrow$} & \makecell{Imaging\\Quality $\uparrow$} & \makecell{Temporal\\Style $\uparrow$}\\
\midrule
Self-Forcing~\cite{huang2025self} & 82.21 & 83.38 & 77.54  & 51.85 & 68.17 & 23.31\\
Infinity-RoPE~\cite{yesiltepe2025infinity} & 83.16 & 84.17 & 79.13  & 57.04 & 67.94 & 23.79\\
DeepForcing~\cite{yi2025deep}  & 82.26 & 83.20 & 78.50 & 52.86 & 67.65 & 23.51\\
MemFlow~\cite{ji2025memflow}     & 82.95 & 83.86 & 79.31 & 60.46 & 67.68 & 23.89\\
LongLive~\cite{yang2025longlive}     & 82.77 & 83.31 & \textbf{80.64} & 40.19 & 68.96 & 23.97\\
\textbf{Anchor Forcing}     & \textbf{83.25} & \textbf{84.40} & 78.67  & \textbf{79.17} & \textbf{69.37} & \textbf{24.17}\\
\bottomrule
\end{tabular}}
\end{table}

\begin{figure}[t]
    \centering
    \includegraphics[width=1.0\linewidth]{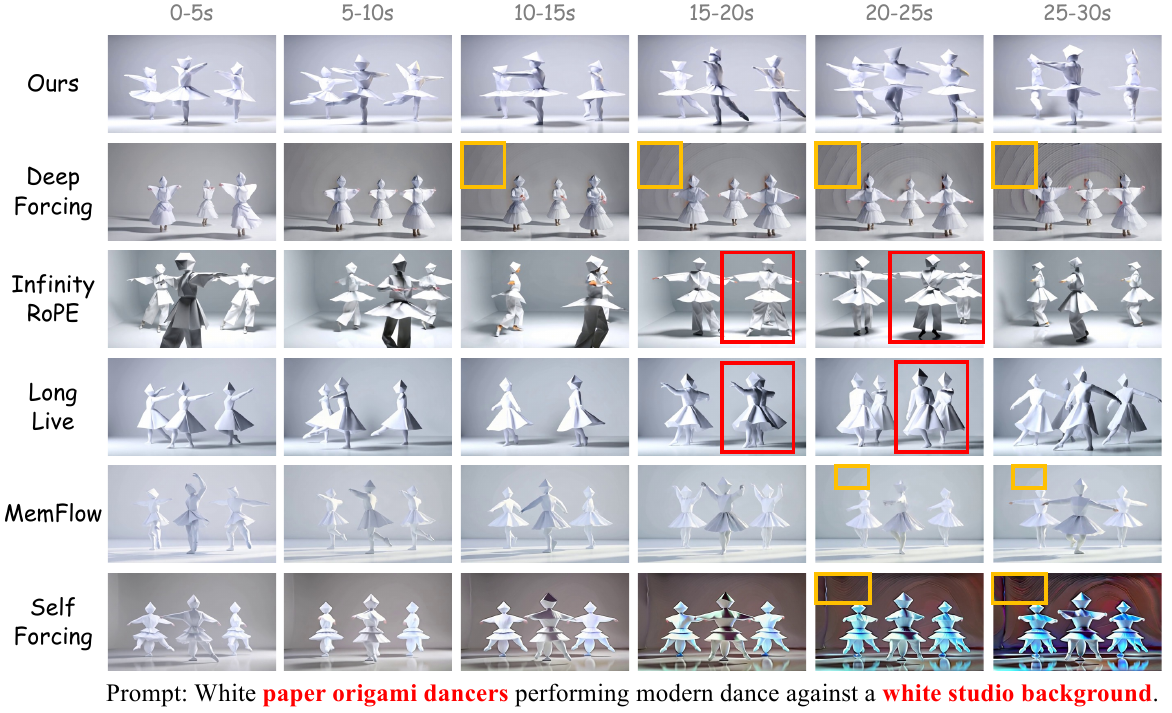}
    \caption{Qualitative comparison on 30-second long-video generation. Compared with prior methods, ours method yields higher-quality, more prompt-faithful videos, avoiding the background and content degradations highlighted in the yellow and red boxes.}
    \label{fig:30s compare.pdf}
\end{figure}

\subsection{Comparisons for Non-Interactive Generation}
To assess the generality of Anchor Forcing beyond prompt interaction, we further compare it with representative non-interactive streaming autoregressive video generation models under single-prompt settings at 5s and 30s. All methods are evaluated on the official VBench prompt set with 946 prompts, using the same resolution of $832\times480$ and the same Wan2.1~\cite{wan2025} backbone.

We compare Anchor Forcing with CausVid~\cite{yin2025causvid}, Self-Forcing~\cite{huang2025self}, Infinity-RoPE~\cite{yesiltepe2025infinity}, DeepForcing~\cite{yi2025deep}, MemFlow~\cite{ji2025memflow}, and LongLive~\cite{yang2025longlive}.
These streaming AR baselines are distilled from Wan-based teachers and share similar model sizes and resolutions.
As shown in~\cref{tab:short}, on the 5-second benchmark, Anchor Forcing achieves the highest overall VBench score among streaming autoregressive models, delivering the best perceptual quality and strong motion performance. Moreover, our method preserves more of the teacher model’s capabilities, particularly in dynamics, achieving performance comparable to the teacher.
Long-horizon single-prompt generation is more challenging due to accumulated distribution shift and compounding temporal errors. We therefore evaluate 30s generation using VBenchLong metrics and report total, quality, semantic, and dynamic scores.
As shown in~\cref{tab:long}, for 30s single-prompt generation, Anchor Forcing ranks first overall and achieves the best total and quality scores. This improvement is consistent across sub-metrics, with the highest imaging quality and temporal style, which indicates stronger long-horizon visual fidelity and stylistic stability. We further provide qualitative results in~\cref{fig:30s compare.pdf}. Compared with prior methods, Anchor Forcing produces higher-quality renderings and follows the prompt more faithfully. In contrast, other baselines exhibit noticeable degradations in either the background or the main content, highlighted by the yellow and red boxes. For example, DeepForcing, MemFlow, and Self-Forcing show progressive background corruption over time, manifesting as ring-like artifacts. Infinity-RoPE and LongLive suffer from content inconsistencies between 15s and 25s, where the number of paper-origami dancers fluctuates across adjacent clips.

\begin{figure}[t]
    \centering
    \includegraphics[width=1.0\linewidth]{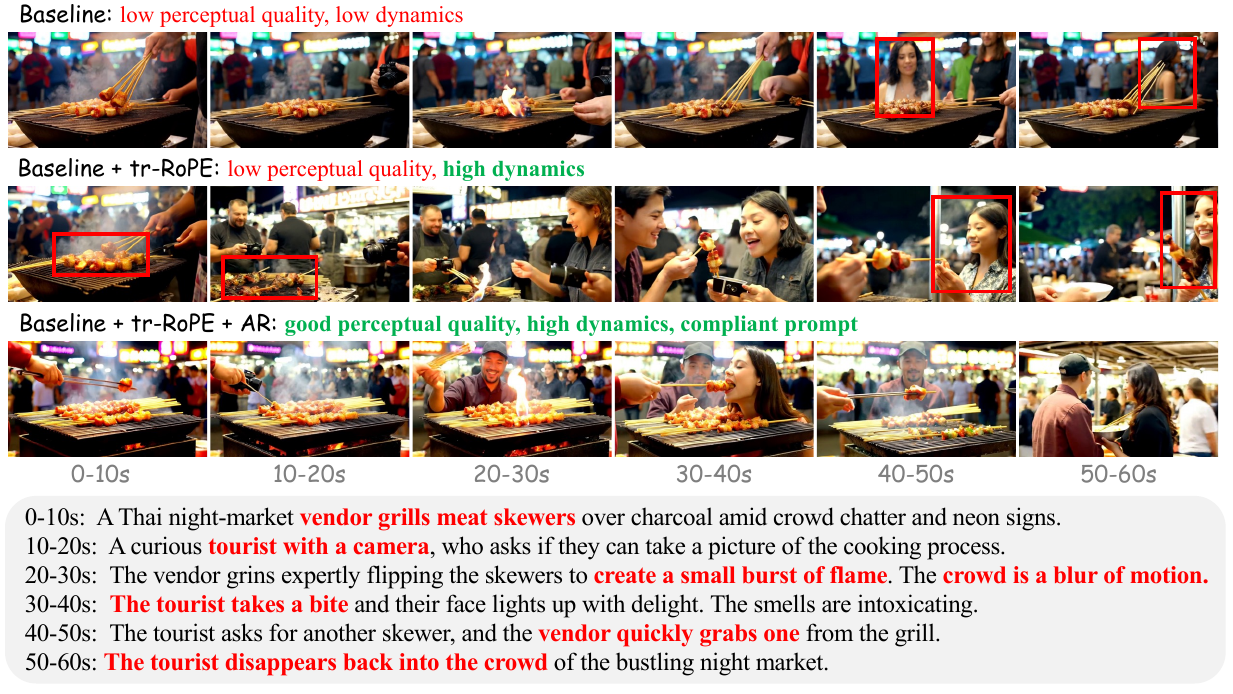}
    \caption{Ablation on 60-second interactive generation. Tri-region RoPE improves motion dynamics, and anchor-guided re-cache further preserves perceptual quality and prompt compliance across segments. Red boxes mark degradation between adjacent clips, and red text denotes newly introduced interactions.}
    \label{fig:ablation}
\end{figure}

\subsection{Ablation Studies}

\begin{wraptable}{r}{0.5\textwidth}
    \centering
    \scriptsize
    \vspace{-12mm}
    \caption{Ablation study on key components. The best results for the module are indicated in bold.}
    \label{tab:ablation}
    \resizebox{\linewidth}{!}{%
    \begin{tabular}{lccc}
    \toprule
    \textbf{Method} & \makecell{CLIP\\Score $\uparrow$} & \makecell{Dynamic\\Degree $\uparrow$} & \makecell{Quality\\Score $\uparrow$} \\
    \midrule
    Baseline  & 23.58 & 26.37 & 78.92 \\
    + tr-RoPE & 23.71 & \textbf{74.83} & 81.72 \\
    + tr-RoPE + AR  & \textbf{23.85} & 73.00 & \textbf{82.25} \\
    \bottomrule
    \end{tabular}}
    \vspace{-5mm}
\end{wraptable}

\textbf{Impact of Tri-region RoPE}. Adding tri-region RoPE (tr-RoPE) to the baseline yields the largest improvement in motion. As shown in \cref{tab:ablation}, the dynamic degree increases from 26.37 to 74.83, indicating that tr-RoPE effectively enhances long-horizon dynamics by using region-wise, range-bounded indexing that keeps RoPE positions within the pretrained regime. We further provide qualitative comparisons in the second row of \cref{fig:ablation}: tr-RoPE produces noticeably more coherent motion than the baseline, but still exhibits appearance degradation and inconsistent prompt responsiveness in some interactive cases. 
For example, in the highlighted regions, the meat skewer details degrade during 0--20s, and the tourist attributes become less faithful during 40--60s. This observation motivates introducing junction caches and anchor-guided re-cache to better preserve boundary evidence and stabilize visual quality across prompt switches.

\noindent\textbf{Impact of Anchor-guided Re-cache Mechanism}. On top of tr-RoPE, introducing the anchor-guided re-cache mechanism (AR) further improves perceptual quality and text alignment. As shown in the last row of \cref{tab:ablation}, adding AR increases the quality score from 81.72 to 82.25 and the CLIP score from 23.71 to 23.85, indicating that AR improves video quality and stabilizes content across prompt switches. We also provide a qualitative comparison in a 60s interactive scenario in \cref{fig:ablation}. Compared with the baseline (first row) and tr-RoPE only (second row), our full model follows the prompts more faithfully: it introduces the new subject in 10--20s, updates the content in 20--30s to ``create a small burst of flame,'' correctly generates the action ``The tourist takes a bite'' in 30--40s, and smoothly transitions the scene in 50--60s to match the prompt ``disappear into the crowd.'' This suggests that adding AR and its anchor memory preserves richer boundary-adjacent evidence, leading to higher perceptual quality while maintaining strong dynamics and achieving the best overall results.

\begin{table}[t]
\centering
\noindent\makebox[\linewidth][c]{%
\begin{minipage}[t]{0.48\linewidth}
\centering
\captionof{table}{Performance of various re-cache.}
\label{tab:repace recache}
\scriptsize
\setlength{\tabcolsep}{3pt}
\renewcommand{\arraystretch}{1.10}
\resizebox{\linewidth}{!}{%
\begin{tabular}{lccc}
\toprule
\textbf{Method} & \makecell{CLIP\\Score $\uparrow$} & \makecell{Dynamic\\Degree $\uparrow$} & \makecell{Quality\\Score $\uparrow$} \\
\midrule
Baseline  & 23.58 & 26.37 & 78.92 \\
Flush   & 23.71 & 27.60 & 79.09 \\
AR   & \textbf{24.22} & \textbf{50.73} & \textbf{79.78} \\
\bottomrule
\end{tabular}}
\end{minipage}%
\hfill
\begin{minipage}[t]{0.48\linewidth}
\centering
\captionof{table}{Performance of various RoPEs.}
\label{tab:repace rope}
\scriptsize
\setlength{\tabcolsep}{3pt}
\renewcommand{\arraystretch}{1.10}
\resizebox{\linewidth}{!}{%
\begin{tabular}{lccc}
\toprule
\textbf{Method} & \makecell{CLIP\\Score $\uparrow$} & \makecell{Dynamic\\Degree $\uparrow$} & \makecell{Quality\\Score $\uparrow$} \\
\midrule
Baseline  & 23.58 & 26.37 & 78.92 \\
Bounded  & 23.58 & 29.09 & 79.38 \\
tr-RoPE     & \textbf{23.71} & \textbf{74.83} & \textbf{81.72} \\
\bottomrule
\end{tabular}}
\end{minipage}
}
\end{table}

\subsection{Analysis}
To further validate the effectiveness of our AR and tr-RoPE, we conduct controlled component replacements on the LongLive baseline~\cite{yang2025longlive}

\noindent\textbf{Replacing re-cache strategies.} As reported in \cref{tab:repace recache}, we replace the original KV re-cache of the baseline with the Flush strategy from Infinity-RoPE~\cite{yesiltepe2025infinity} and our AR, while keeping other settings unchanged.
Flush yields only marginal improvements over re-cache, as reflected by the CLIP score increasing from 23.58 to 23.71 and the dynamic degree rising from 26.37 to 27.60, whereas our AR achieves the best performance across all metrics with a CLIP score of 24.22, a dynamic degree of 50.73, and a quality score of 79.78. 
This gain is attributed to AR explicitly leveraging anchor memory that integrates sink, clip, and local caches, which jointly preserve global semantics, clip evidence, and recent context for post-switch conditioning, thereby improving prompt adherence and perceptual quality.

\noindent\textbf{Replacing RoPE variants.} We further compare three positional encoding variants on the same baseline: Basic RoPE with unbounded indexing, Bounded RoPE that follows the pretrained positional limit (e.g., capped at 21), and our tri-region RoPE.
As shown in \cref{tab:repace rope}, bounded RoPE provides only a modest improvement in dynamics and quality, while tri-region RoPE delivers a substantially larger gain, especially in dynamic degree (26.37$\rightarrow$74.83), together with improved CLIP score (23.58$\rightarrow$23.71) and quality score (78.92$\rightarrow$81.72).
These results indicate that tri-region RoPE is critical for retaining motion priors under long-horizon streaming, and it complements AR by improving motion dynamics while AR stabilizes boundary conditioning and content fidelity.

\section{Conclusion}
In this work, we propose \textbf{Anchor Forcing}, a cache-centric framework for interactive long video generation under streaming constraints. Anchor Forcing combines an anchor-guided re-cache mechanism, which builds an anchor memory from sink, junction, and local caches to stabilize post-switch conditioning, with a tri-region RoPE that uses region-specific, range-bounded indexing to better retain pretrained motion priors over long horizons. Extensive experiments on VBench show that Anchor Forcing improves perceptual quality and motion dynamics in interactive settings and consistently outperforms prior streaming baselines, achieving state-of-the-art results.


%
%
\bibliographystyle{splncs04}
\bibliography{main}
\end{document}